\documentclass[10pt,twocolumn,letterpaper]{article}

\usepackage[pagenumbers]{iccv} 

\definecolor{iccvblue}{rgb}{0.21,0.49,0.74}
\usepackage[pagebackref,breaklinks,colorlinks,allcolors=iccvblue]{hyperref}

\usepackage{multicol}
\usepackage{multirow}
\usepackage{amsmath}
\usepackage{bm}
\usepackage{amsmath, amssymb, amsthm} 

\theoremstyle{plain}
\newtheorem{theorem}{Theorem}[section]

\theoremstyle{definition}
\newtheorem{definition}[theorem]{Definition}

\theoremstyle{remark}
\newtheorem{remark}[theorem]{Remark}

\def\paperID{1066} 
\def\confName{ICCV}
\def\confYear{2025}

\title{\textit{ForgetMe}: Evaluating Selective Forgetting in Generative Models}

\author{
Zhenyu Yu\\
Universiti Malaya\\
{\tt\small yuzhenyuyxl@foxmail.com}
\and
Mohd Yamani Inda Idris\\
Universiti Malaya\\
{\tt\small yamani@um.edu.my}
\and
Pei Wang\\
Kunming University of \\Science and Technology\\
{\tt\small peiwang@kust.edu.cn}
}

\begin{document}

\twocolumn[{
\renewcommand\twocolumn[1][]{#1}
\maketitle
\begin{center}
    \captionsetup{type=figure}
    \includegraphics[width=1.0\linewidth]{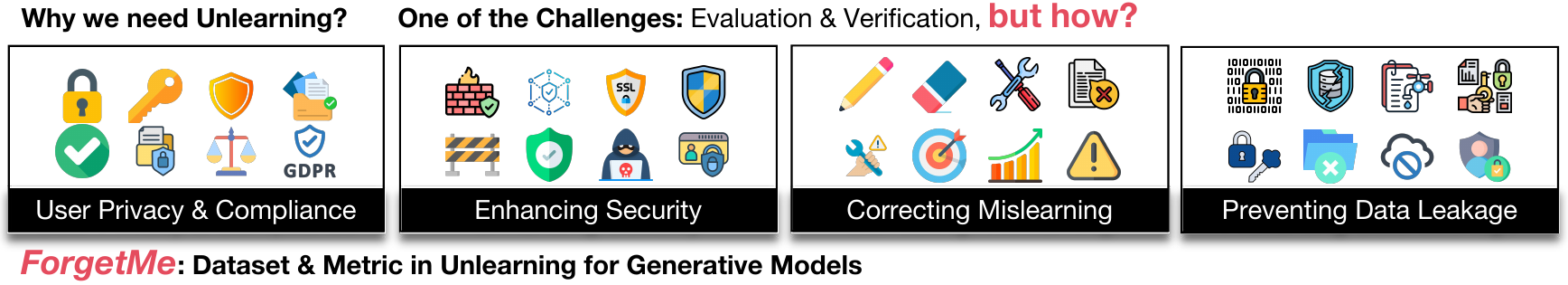} 
    \caption{Motivation for proposing \textit{ForgetMe}: (1) Why we need models capable of unlearning, especially for privacy compliance and ethical considerations; (2) A core challenge in unlearning—how to effectively evaluate and verify the unlearning process; (3) We proposed the \textit{ForgetMe}, providing a dedicated dataset and metric specifically for evaluating selective forgetting in generative models.}
    \label{fig_motivation}
\end{center}
}]

\begin{abstract}
The widespread adoption of diffusion models in image generation has increased the demand for privacy-compliant unlearning. However, due to the high-dimensional nature and complex feature representations of diffusion models, achieving selective unlearning remains challenging, as existing methods struggle to remove sensitive information while preserving the consistency of non-sensitive regions. To address this, we propose an Automatic Dataset Creation Framework based on prompt-based layered editing and training-free local feature removal, constructing the \textit{ForgetMe} dataset and introducing the \textit{Entangled} evaluation metric. The \textit{Entangled} metric quantifies unlearning effectiveness by assessing the similarity and consistency between the target and background regions and supports both paired (\textit{Entangled-D}) and unpaired (\textit{Entangled-S}) image data, enabling unsupervised evaluation. The \textit{ForgetMe} dataset encompasses a diverse set of real and synthetic scenarios, including CUB-200-2011 (Birds), Stanford-Dogs, ImageNet, and a synthetic cat dataset. We apply LoRA fine-tuning on Stable Diffusion to achieve selective unlearning on this dataset and validate the effectiveness of both the \textit{ForgetMe} dataset and the \textit{Entangled} metric, establishing them as benchmarks for selective unlearning. The dataset and code will be publicly released upon paper acceptance. Our work provides a scalable and adaptable solution for advancing privacy-preserving generative AI.

\end{abstract}



\section{Introduction}
Existing research on model unlearning has focused on computer vision tasks like object removal and scene modification \cite{liang2024cmat, wang2024enhancing,yi2025score,xin2023self,yi2024towards} and natural language processing \cite{zhou2024human, yang2024wcdt,wang2025mdanet}, primarily relying on fine-tuning or pruning. For generative models, \textbf{feature forgetting} \cite{xin2025resurrect,zhou2025glimpse,xin2024vmt,xin2024mmap,qin2025lumina} is crucial for precise removal of specific features while maintaining controllability. However, current methods struggle to balance forgetting accuracy and background preservation, often degrading generation consistency and affecting non-forgotten regions.

\textbf{Need for a New Dataset.}
Most existing unlearning datasets rely on \textit{data poisoning} techniques (e.g., BadNets \cite{ye2023mplug,zhou2023analyzing,wang2023evaluation,zhou2024calibrated,zhou2025anyprefer,xin2025lumina,xin2024parameter,xin2024v}) by injecting backdoors or noise into training data. However, these methods have limited applicability to generative and multimodal models due to their high-dimensional nature, where a single sample can influence the entire data distribution. Thus, a {standardized dataset} covering diverse modalities (real and synthetic images) and tasks is essential for robust unlearning evaluation.

\textbf{Need for a New Evaluation Metric.}
While current unlearning methods effectively remove target features, they often fail to maintain {background consistency}, leading to semantic instability. Existing metrics focus on data removal but overlook {retention consistency}, which assesses the preservation of non-forgotten regions. A unified evaluation metric is needed to measure both forgetting effectiveness and stability, ensuring reliable and practical model unlearning.

To address these challenges, we propose the \textit{ForgetMe}, which includes a dedicated evaluation metric and dataset to quantify selective unlearning performance in generative models. Our \textbf{contributions} are as follows:


\begin{itemize}
    \item \textbf{Entangled Evaluation Metric:} We propose a novel evaluation metric, \textit{Entangled}, designed to assess unlearning effectiveness across various image generation tasks. The metric supports both paired and unpaired image data, allowing precise quantification of a model's \textit{forgetting} capability.
    
    \item \textbf{Automatic Dataset Creation Framework:} Based on the concept of layered image editing, we utilize existing pre-trained models and construct a training-free removal framework, using Entangled as evaluation metrics.
    
    \item \textbf{ForgetMe Dataset:} We release the dataset based on a training-free removal framework, which includes a diverse range of real-world and synthetic images to support comprehensive multi-task unlearning evaluation for generative models.

\end{itemize}

\begin{figure*}[!ht]
    \centering
    \includegraphics[width=1.0\linewidth]{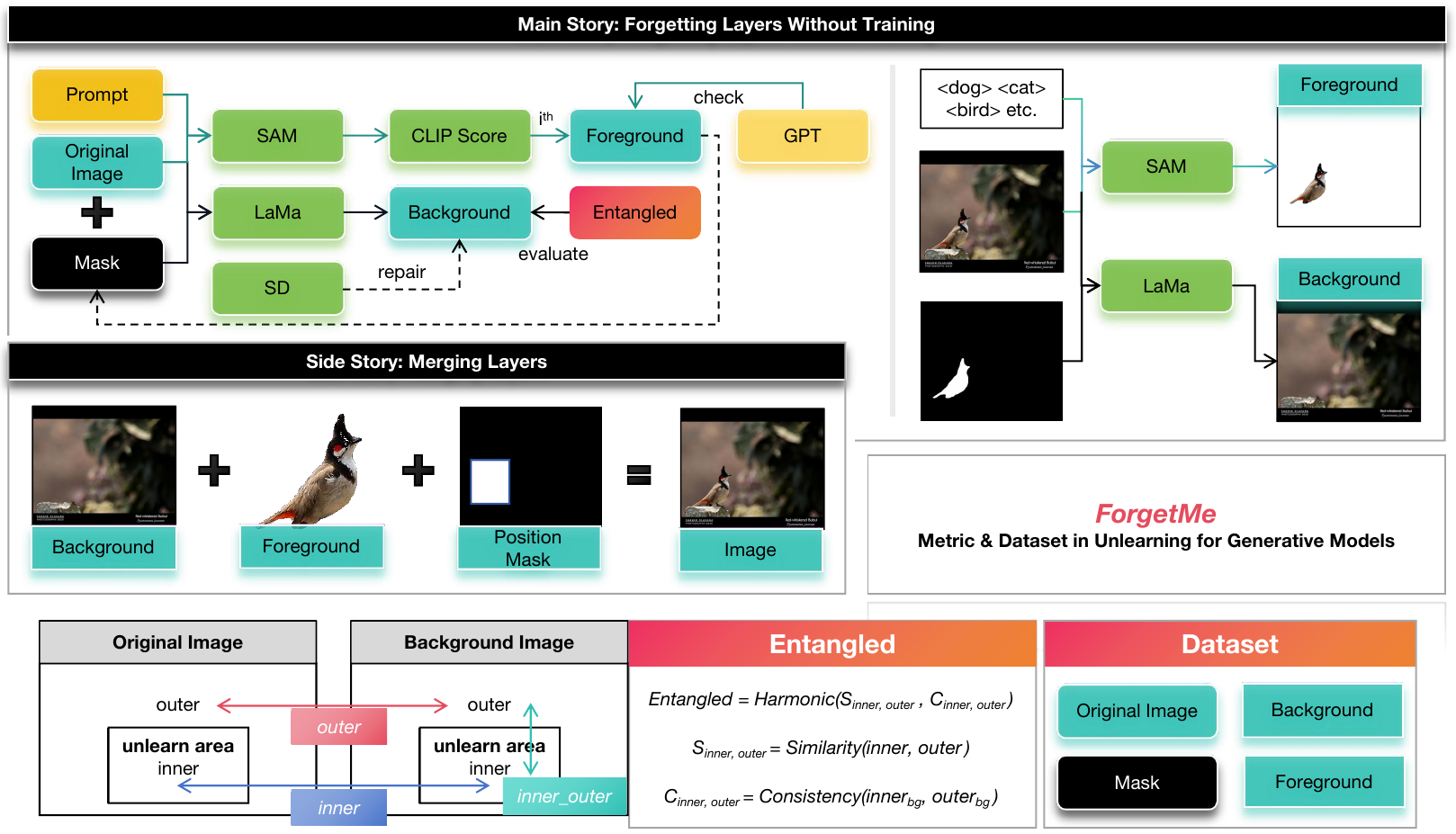}
    \caption{Architecture of the proposed Automatic Dataset Creation Framework, which is based on {prompt-based layered editing} and {training-free local feature removal}. The framework consists of three core components: \textbf{Main Story}, which is designed for dataset construction from natural scene images by performing {Foreground Extraction} to isolate target information and {Background Reconstruction} to restore structural integrity, forming the \textit{ForgetMe Dataset}; \textbf{Side Story}, which is tailored for synthesizing datasets using pre-existing foreground and background images, incorporating the {Merging Layers} module to enhance structural consistency and enable reversible forgetting; and the \textit{Entangled Metric}, which quantifies removal effectiveness by measuring residual associations between foreground and background.}
    \label{fig_architecture}
\end{figure*}

\section{Related Work}

\subsection{Model Unlearning}
\textbf{Model Unlearning} is defined as the process of making a model \textit{forget} specific data or concepts without fully retraining the model. This capability is essential for meeting privacy requirements, such as the GDPR's right to be forgotten, ensuring that models do not retain sensitive information after it is deleted from training data \cite{liu2024safer, chen2024machine,yu2025yuan}. The primary objective of model unlearning is to effectively remove the influence of specific data on the model’s parameters without compromising overall performance \cite{bommasani2021opportunities, xu2024survey,yu2025qrs}.

Existing unlearning methods include fine-tuning, incremental updates, parameter pruning, and adversarial training, primarily applied in computer vision and natural language processing (NLP) tasks to remove specific data or features from models \cite{yao2023llm_unlearning, wu2024deep_unlearning,yu2025guideline}. However, these methods typically require substantial parameter adjustment or retraining, especially for large-scale models, resulting in high computational costs and resource demands \cite{chen2024machine,yu2025ai}. Additionally, these methods often struggle to preserve generation quality while achieving target-specific unlearning, especially in high-dimensional generative models \cite{liu2024safer}.

\subsection{Diffusion Model Unlearning}
\textbf{Concept Forget} refers to the selective removal of specific concepts or features (e.g., facial attributes or style elements) from diffusion models to ensure they no longer appear in generated images. Gandikota et al. \cite{gandikota2023inpaint_unlearn, chen2024score} proposed a method for editing the latent space of diffusion models to achieve concept forgetting without retraining. However, while removing target concepts is feasible, it often leads to significant alterations in the overall image content, disrupting structural consistency and visual coherence, especially in high-dimensional diffusion models \cite{suriyakumar2024unstable, zhu2024choose}.

\textbf{Object Removal} in diffusion models requires the deletion of specific objects or regions and seamless inpainting to maintain natural visual coherence. Inst-Inpaint \cite{yildirim2023inst, chen2024fast} leverages diffusion models for object removal, using masks and text prompts to control the removal process. However, existing approaches struggle with consistency in complex scenes, with edge inconsistencies and texture mismatches frequently visible in the inpainted regions, particularly for high-resolution images \cite{gao2024meta, gandikota2023inpaint_unlearn}.

\subsection{Model Unlearning vs. Object Removal}
{Model Unlearning} and {Object Removal} both aim to eliminate specific content from generated outputs, but their core objectives and methodologies differ significantly. \textbf{Model Unlearning} focuses on modifying the learned representations within a model to ensure that certain information is no longer accessible or influential during generation. This process typically involves techniques such as {fine-tuning}, {gradient-based updates}, or {explicit data deletion} \cite{fan2023salun, heng2024selective, lu2024mace,he2025enhancing,wang2025see}. It is primarily applied in privacy compliance scenarios, such as meeting GDPR requirements, ensuring that models do not retain specific training data or learned concepts. However, since model unlearning directly alters model parameters, it is challenging to precisely locate and remove specific data without affecting the overall performance of the model. The {uncontrollability of forgetting} remains a major challenge in this field.

In contrast, \textbf{Object Removal} operates at the {instance level} by utilizing {mask-based inpainting} methods to erase specific objects while preserving the surrounding background integrity \cite{yildirim2023inst, chen2024fast,he2024ddpm,xiang2025regrap,wang2025twin}. These methods typically rely on the image inpainting capabilities of diffusion models, such as Inst-Inpaint, for object removal. 
However, existing approaches still face challenges related to consistency and manual intervention, as inpainted regions in complex scenes often suffer from edge discontinuities and texture mismatches, affecting visual coherence, especially in high-resolution images \cite{gao2024meta,yu2025dancetext,yu2025dc4cr}. Additionally, many object removal methods rely on manually provided masks or textual prompts, restricting automation and increasing user effort.

\subsection{Challenges in Unlearning}
\textbf{Evaluation Metrics.} Current evaluation metrics for generative models, such as Frechet Inception Distance (FID) \cite{chong2020effectively, liu2024survey, mcwilliam2006learning,yu2025physics2}, effectively measure the quality of generated images but fail to fully capture unlearning performance. Other CLIP-based metrics, including CLIP Distance and CLIP Maximum Mean Discrepancy (CMMD) \cite{jayasumana2024rethinking, yildirim2023inst,yu2025physics}, assess semantic alignment in generated images but are limited in evaluating inner-outer region consistency and specific target removal effectiveness. The high dimensionality and data diversity of diffusion models add complexity to establishing unified evaluation standards \cite{tian2024practical_unlearning, xu2024machine,sarkar2025reasoning}.

\textbf{Datasets.} Few public datasets are designed for model unlearning evaluation, and traditional datasets like ImageNet \cite{deng2009imagenet,li2025multi} and COCO \cite{lin2014microsoft,zhou2025reagent,wang2025unitmge} lack task diversity and complexity, making them inadequate for assessing selective unlearning in diffusion models \cite{gao2024practical_unlearning, huang2024offset_unlearning,liang2024self,li2024falcon}. Gandikota et al. \cite{gandikota2023inpaint_unlearn} emphasized the need for high-quality, task-diverse synthetic datasets for evaluating concept forgetting and object removal. However, real-world datasets like {RORD} \cite{sagong2022rord}, while useful for object removal, lack category labels, limiting their suitability for prompt-based feature unlearning tasks that require well-defined target concepts.

\begin{figure*}
    \centering
    \includegraphics[width=1\textwidth]{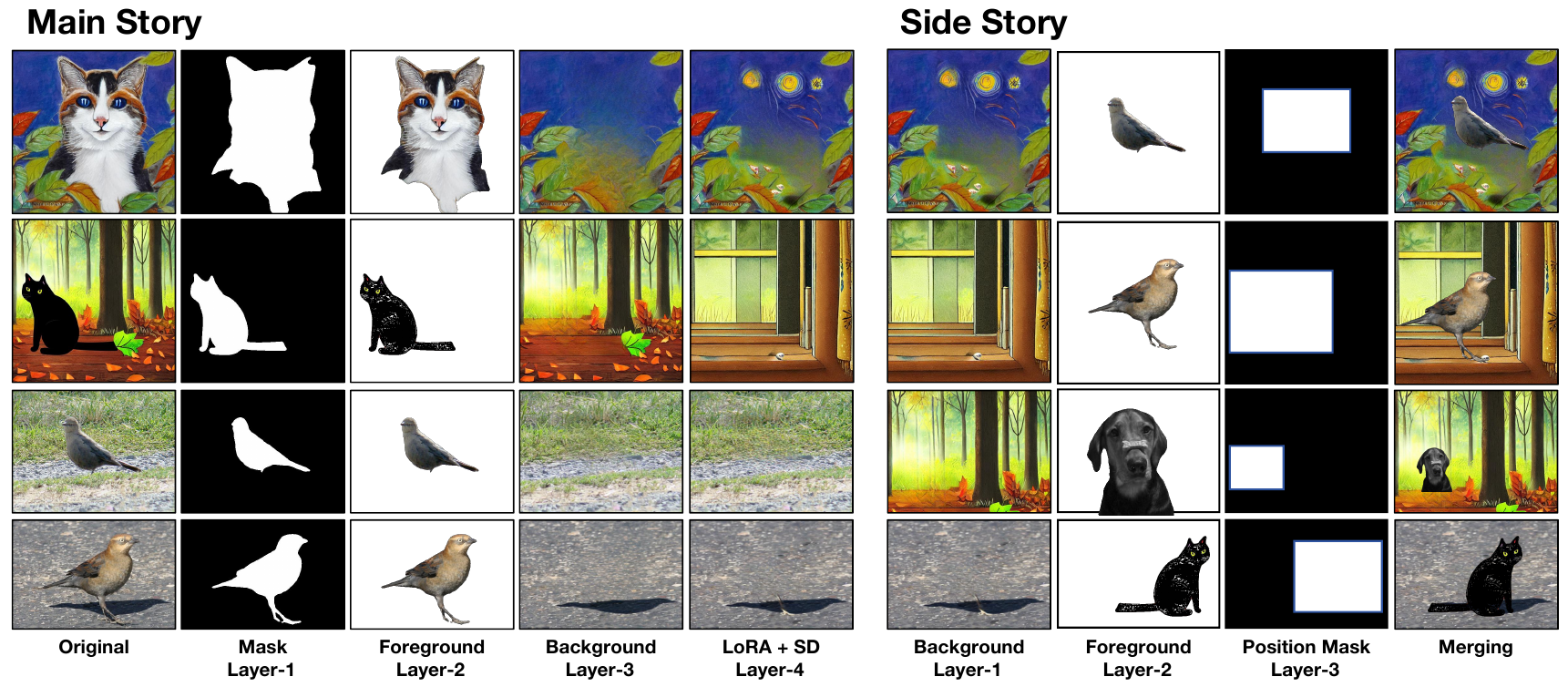}
    \caption{The visualization of main story and side story perspectives. The main story illustrates the layered decomposition of the original image into foreground, background, and mask layers, where the background layer represents the original image with the foreground effectively unlearned. The side story combines the foreground and background layers to showcase the reconstructed image, demonstrating the model's ability to achieve seamless unlearning and reintegration of image components. It is noteworthy that all of this requires no additional training.}
    \label{fig_results}
\end{figure*}

\section{Method}
We propose an automatic dataset creation framework that enables the precise removal of specific information while maintaining visual coherence (see Figure \ref{fig_architecture}). Based on this framework, the \textit{ForgetMe} dataset is generated to provide data support for selective unlearning. The framework consists of the following three key components:




(1) \textbf{Main Story}. The \textit{ForgetMe} dataset construction is designed for natural scene images and consists of three key steps: (a) {Foreground Extraction}, which isolates target information, and (b) {Background Reconstruction}, which restores structural integrity.

(2) \textbf{Side Story (Optional)}. This optional component is tailored for dataset synthesis using pre-existing foreground and background images. It incorporates the {Merging Layers} module to recombine foreground, background, and masks, enhancing structural consistency and enabling reversible forgetting.

(3) \textbf{Entangled Metric}. We propose the \textit{Entangled} metric to quantify removal effectiveness by measuring residual associations between the foreground and background, ensuring a balanced evaluation of forgetting and retention.

\subsection{Main Story: Forgetting Layers}
\subsubsection{Foreground Extract}

The goal of foreground extraction is to isolate the specific object or information that needs to be removed from the image. The steps for this process are as follows:

\textit{Step 1: Generate segmentation masks.} SAM (Segment Anything Model) \cite{kirillov2023segment} is used to produce multiple segmentation masks on the image, dividing it into various regions.

\textit{Step 2: Calculate CLIP scores for each mask.} Using the CLIP model, a similarity score (CLIP score \cite{hessel2021clipscore}) is computed for each mask, assessing how closely each region corresponds to the target information.

\textit{Step 3: Select the highest-scoring mask and crop the foreground region.} The mask with the highest CLIP score is selected, and the corresponding region is cropped as an initial foreground area.

\textit{Step 4: Validate the foreground using GPT.} The cropped foreground image is input to GPT along with a verification prompt.  

\textit{Q: Is this a [target category]?} 

\textit{A: Yes, this is a [target category]. }

\textit{Solution: The extraction is considered successful.}

\textit{A: No, this is not a [target category]. }

\textit{Solution: We return to Step 3 and select the next highest-scoring mask for foreground extraction. }

\textit{Step 5: Final foreground confirmation.} The process repeats until GPT successfully verifies the foreground as belonging to the intended category, designating it as the final foreground selection.







\subsubsection{Background Reconstruction}

The goal of background reconstruction is to restore the missing region left by the removed foreground, ensuring visual coherence and structural consistency. The specific steps are as follows:

\textit{Step 1: Prepare the image for inpainting.} The region corresponding to the removed foreground is masked out, creating an incomplete image with a blank area for inpainting.

\textit{Step 2: Initial inpainting with LaMa.} The LaMa model \cite{suvorov2022resolution} is applied to the blank region, performing an initial inpainting to generate a plausible background.

\textit{Step 3: Evaluate background quality.} The \textit{Entangled} metric is used to assess the reconstructed background by measuring its residual association with the removed foreground. If the score is low, indicating poor reconstruction quality, further refinement is required.

\textit{Step 4: Refine prompt and reapply inpainting.} For backgrounds with low \textit{Entangled} scores, the inpainting prompt is manually refined to enhance the generation process. Stable Diffusion (SD) \cite{rombach2022high} is then employed for a second inpainting pass to enhance background coherence and realism.

\subsection{Side Story: Merging Layers}

The merging layers process consists of the following steps:

\textit{Step 1: Align foreground and background using the position mask.} The position mask serves as a spatial reference to accurately position the foreground over the background. This step ensures that the foreground is placed correctly, preserving the original spatial structure and maintaining the natural composition of the image.

\textit{Step 2: Seamlessly blend the layers.} Once aligned, the foreground and background layers are blended at the edges to minimize any visible seams or inconsistencies. This blending process smooths transitions between layers, reducing artifacts and ensuring a visually coherent result.

\textit{Step 3: Generate the final merged image.} After alignment and blending, the layers are combined into a single, cohesive image. The final output maintains the structural integrity of the original scene while ensuring that the removed content does not introduce visual disruptions or inconsistencies.

\subsection{Evaluation Metric - \textbf{\textit{Entangled}}}


In removal tasks, our goal is to make the removed regions as distinct as possible, while retaining regions that remain as consistent as possible. This application scenario inspired us to propose a quantitative evaluation metric for selective unlearning, called \textit{Entangled}. This metric assesses the residual association between the foreground and background in the modified images, measuring the thoroughness of the unlearning process. It combines similarity and consistency metrics to ensure maximum separation between foreground and background information after unlearning.

\begin{definition}[Entangled Metric]
The \textit{Entangled} metric quantifies similarity and consistency between inner and outer regions of original and background images. We define:
\begin{equation}
\begin{split}
    Entangled &= \frac{\alpha + \beta}{\frac{\alpha}{S_{\text{inner, outer}}} + \frac{\beta}{C_{\text{inner, outer}}}} \\
    &= \frac{(\alpha + \beta) \cdot S_{\text{inner, outer}} \cdot C_{\text{inner, outer}}}{\alpha \cdot C_{\text{inner, outer}} + \beta \cdot S_{\text{inner, outer}}}
\end{split}
\end{equation}
where \( S_{\text{inner, outer}} \) is the combined similarity, and \( C_{\text{inner, outer}} \) is the consistency metric, with weights \(\alpha, \beta \geq 0\) such that:
$\alpha + \beta = 1$.
Adjusting \(\alpha\) and \(\beta\) prioritizes either similarity or consistency; setting \(\alpha = 0\) and \(\beta = 1\) emphasizes consistency in single-image scenarios. The default \(\alpha = \beta = 0.5\) balances both aspects.
\end{definition}

\begin{definition}[Similarity]
Let \( X \) and \( Y \) be pixel matrices for original and background images, normalized to \( X, Y \in [0, 1] \). Define regions \( \Omega_{\text{inner}} \) and \( \Omega_{\text{outer}} \) with pixels \( X_i, Y_i \in \Omega_{\text{inner}} \) and \( X_j, Y_j \in \Omega_{\text{outer}} \). 



\begin{equation}
    S_{\text{region}} = \sqrt{\frac{1}{N_{\text{region}}} \sum_{k \in \Omega_{\text{region}}} (X_k - Y_k)^2}
\end{equation}
where $N_{region} \in N_{inner}, N_{outer}$, ${k} \in {i}, {j}$, and \( N \) is total pixel counts.

The rationale for selecting this function is discussed in Section \ref{statistic_questionnaire}. To ensure consistency with a monotonically increasing relationship, regardless of proximity, the outer similarity is adjusted as \( 1 - S_{\text{outer}} \). The combined similarity metric \( S_{\text{inner, outer}} \) is defined as:
\begin{equation}
S_{\text{inner, outer}} = \frac{2 \cdot S_{\text{inner}} \cdot (1 - S_{\text{outer}})}{S_{\text{inner}} + (1 - S_{\text{outer}}) + \epsilon}
\end{equation}
where \(\epsilon = 1 \times 10^{-6}\) prevents division by zero.
\end{definition}

\begin{remark}
The combined similarity metric \( S_{\text{inner, outer}} \) attains its maximum value of 1, i.e., 
\[
S_{\text{inner, outer}} \to 1 \quad \text{as} \quad S_{\text{inner}} \to 1, \, S_{\text{outer}} \to 0.
\]
Conversely, \( S_{\text{inner, outer}} \) reaches its minimum value of 0, i.e.,
\[
S_{\text{inner, outer}} \to 0 \quad \text{as} \quad S_{\text{inner}} \to 0, \, S_{\text{outer}} \to 1.
\]
\end{remark}

\begin{definition}[Consistency]
The consistency metric \( C_{\text{inner, outer}} \) measures alignment of statistical characteristics (mean and variance) between inner and outer regions to maintain coherence. Define:
\begin{equation}
    C_{\text{inner, outer}} = M \times V
\end{equation}
where mean consistency \( M \) and variance consistency \( V \) are given by:

\begin{equation}
    M = \frac{2 \cdot \mu_{\text{inner}} \cdot \mu_{\text{outer}}}{\mu_{\text{inner}}^2 + \mu_{\text{outer}}^2 + \epsilon}
\end{equation}
\begin{equation}
    V = \frac{2 \cdot \sigma_{\text{inner}} \cdot \sigma_{\text{outer}}}{\sigma_{\text{inner}}^2 + \sigma_{\text{outer}}^2 + \epsilon}
\end{equation}
where:
\begin{equation}
    \mu_{\text{region}} = \frac{1}{N_{\text{region}}} \sum_{k \in \Omega_{\text{region}}} X_k 
\end{equation}

\begin{equation}
    \sigma_{\text{region}}^2 = \frac{1}{N_{\text{region}}} \sum_{k \in \Omega_{\text{region}}} (X_k - \mu_{\text{region}})^2
\end{equation}
where \( \epsilon = 1 \times 10^{-6} \) prevents division by zero.
\end{definition}

\begin{remark}
The consistency metric \( C_{\text{inner, outer}} \) approaches its maximum value of 1, i.e.,
\begin{align}
    C_{\text{inner, outer}} \to 1 \quad 
    &\text{as} \quad \left| \mu_{\text{inner}} - \mu_{\text{outer}}\right| \to 0, \, \nonumber \\
    &\text{and} \quad \left| \sigma_{\text{inner}} - \sigma_{\text{outer}} \right| \to 0. \nonumber
\end{align}
Higher \( C_{\text{inner, outer}} \) indicate a closer match in statistical characteristics, promoting visual coherence. Conversely,
\begin{align}    
    C_{\text{inner, outer}} \to 0 \quad 
    &\text{as} \quad \left| \mu_{\text{inner}} - \mu_{\text{outer}} \right| \to +\infty, \quad \nonumber \\
    &\text{or} \quad \left| \sigma_{\text{inner}} - \sigma_{\text{outer}} \right| \to +\infty \nonumber,
\end{align}
where the metric approaches its minimum value of 0, suggesting greater statistical discrepancy and potentially incomplete background preservation.
\end{remark}

\begin{figure*}
    \centering
    \includegraphics[width=1\linewidth]{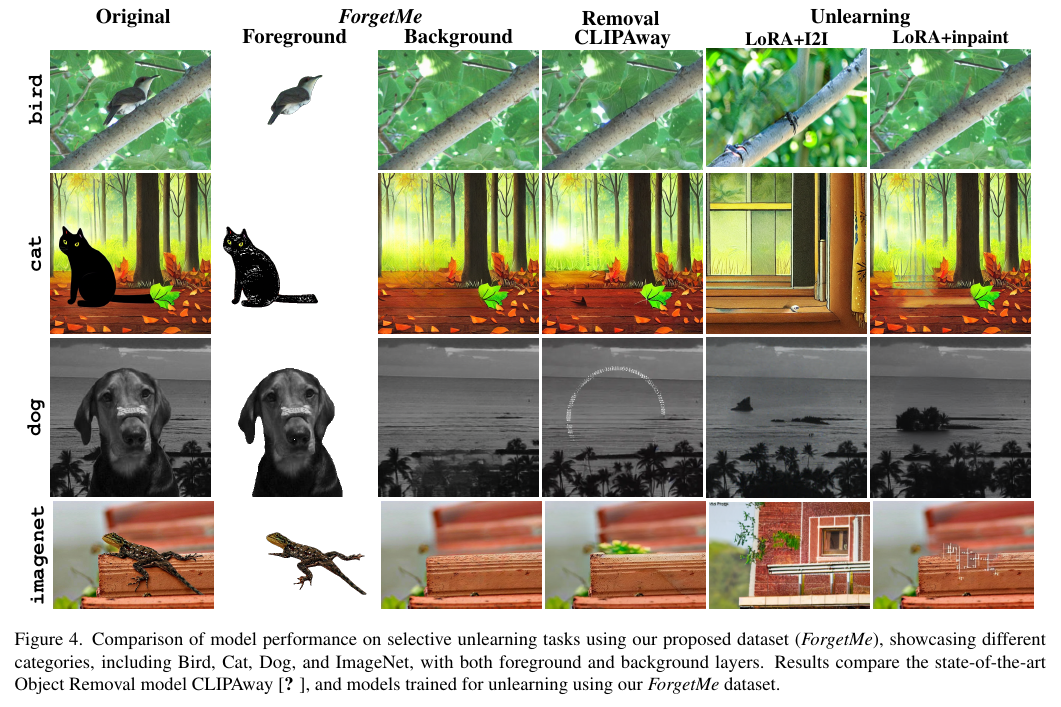}
    \label{fig_compare_models}
\end{figure*}

\begin{figure*}
    \centering
    \includegraphics[width=1\linewidth]{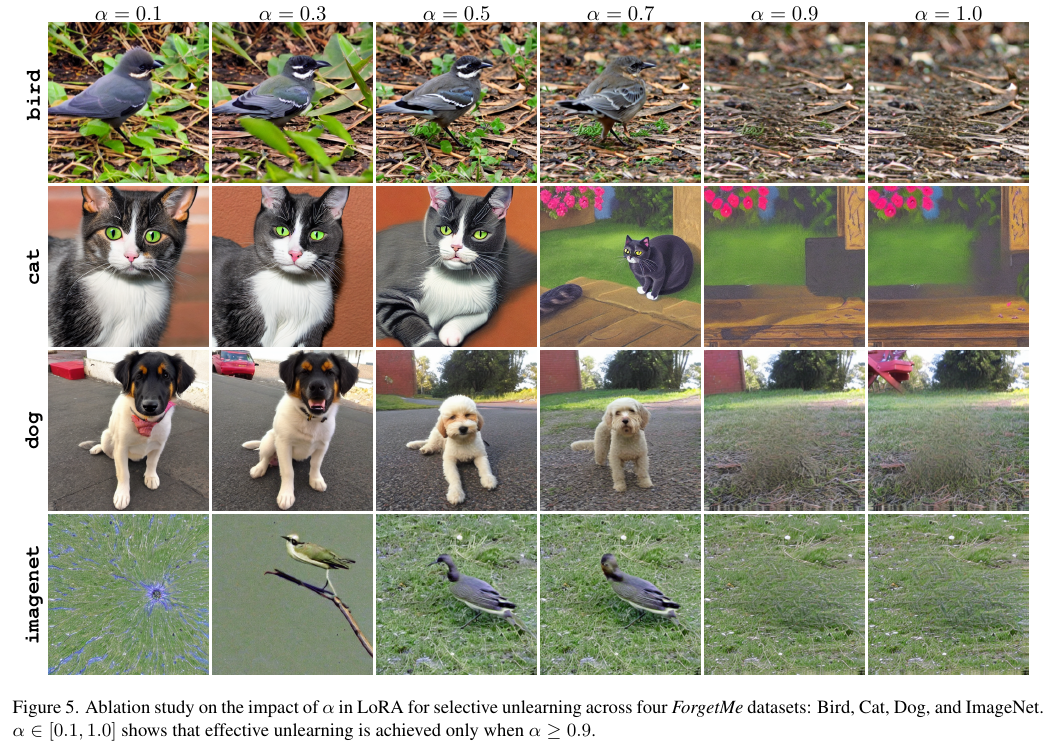}
    \label{fig_ablation_lora_alpha}
\end{figure*}

\section{Experiment}
\subsection{Dataset Description}
We selected three public datasets and one synthetic dataset to evaluate our unlearning framework.
\textbf{ImageNet} \cite{deng2009imagenet}: The validation set was used, with prompts derived from class labels for contextual unlearning.
\textbf{Stanford Dogs} \cite{khosla2011novel}: Prompts were standardized as \textit{dog} to focus on generic dog breed representations.
\textbf{CUB-200-2011 (Bird)} \cite{wah2011caltech}: The prompt \textit{bird} was used to assess unlearning across diverse bird species.
\textbf{Generated-cats}: A synthetic dataset designed for evaluating unlearning on generated cat images. Using GPT-4o, 10,000 cat-related prompts were generated and used with Stable Diffusion V1.5 to create $512 \times 512$ images, each paired with the prompt \textit{cat}.
The combination of real and synthetic datasets provides a diverse evaluation framework, ensuring comprehensive testing of the model’s generalizability and robustness in unlearning tasks.

\subsection{Experimental Settings}
Experiments were conducted on a single NVIDIA A100 GPU with 80 GB memory, ensuring sufficient computational capacity for high-resolution image processing and large-scale unlearning tasks. All models operated with default settings to maintain consistency. Specifically, SAM was used for segmentation, and LaMa for background inpainting, both without additional fine-tuning. GPT-4o assisted in content validation.

For fine-tuning Stable Diffusion V1.5’s U-Net, we applied LoRA \cite{hu2021lora} with prompts \texttt{<cat>}, \texttt{<dog>}, \texttt{<bird>}, and \texttt{<imagenet>}. To ensure consistency, \texttt{<imagenet>} was used across all ImageNet data. The model was trained at a $512 \times 512$ resolution with a batch size of 1, using a learning rate of $1e^{-4}$ over 30,000 iterations, leveraging \textit{ForgetMe} background images. The dataset consists of four components: background, foreground, original image, and mask. Foreground deletion success rates exceeded 90\% (see Table \ref{table_datasets_info}), demonstrating the effectiveness of this removal framework.

\subsection{Evaluation Metrics}
\textbf{Frechet Inception Distance (FID)} \cite{chong2020effectively} evaluates image quality by comparing statistical features of real and generated images using a pretrained Inception network. 
\textbf{CLIP Maximum Mean Discrepancy (CMMD)} \cite{jayasumana2024rethinking} quantifies distributional alignment between original and inpainted images using CLIP features. 
\textbf{CLIP Distance} \cite{yildirim2023inst} measures the cosine distance between CLIP embeddings of original and inpainted images.
\textbf{CLIP Accuracy} \cite{ekin2024clipaway}, adapted from CLIPAway, tracks class changes post-unlearning using CLIP as a zero-shot classifier. Success rates at Top-1, Top-3, and Top-5 (CLIP@1, CLIP@3, CLIP@5) indicate unlearning effectiveness.
\textbf{\textit{Entangled}} assesses removal by measuring similarity and consistency between inner (target) and outer (background) regions. 


\subsection{ForgetMe Dataset}




Our framework demonstrates a structured process for selective removal without additional training (see Figure \ref{fig_results}). In the \textbf{main story}, the original image is decomposed into layers, including foreground, background, and mask. The background layer effectively represents the original image with the foreground removed, achieving a visually coherent removal effect. This layered structure ensures the removal of target information while preserving the background's integrity.
In the \textbf{side story}, the foreground and background layers are recombined through a position mask. This merging of layers maintains visual quality and consistency, serving as an exploration of layer merging rather than a primary focus of this work. This ability to separate and recombine image layers without retraining highlights the flexibility and adaptability of our proposed framework.

\begin{table}[!ht]
    \centering
    \caption{Summary of Dataset Information. It provides an overview of the datasets we used and generated. This structured dataset setup helps us evaluate the effectiveness of the unlearning framework across various scenarios, from real-world images to synthetically generated content.}
    \resizebox{1.0\linewidth}{!}{
    \begin{tabular}{cccccc}
    \toprule
        \textbf{Dataset} & \textbf{Prompt} & \textbf{Images} & \textbf{Selected} & \textbf{Success} \\ \midrule
        Bird & \texttt{$<$bird$>$} & 11,788 & 11,486 & 97.44\% \\ 
        Cat & \texttt{$<$cat$>$} & 10,000 & 9,555 & 95.55\% \\ 
        Dog & \texttt{$<$dog$>$} & 20,258 & 18,408 & 90.87\% \\ 
        ImageNet & \texttt{$<$imagenet$>$} & 5,000 & 4,377 & 87.54\% \\ \bottomrule
    \end{tabular}
    }
    \label{table_datasets_info}
\end{table}

\begin{table*}[!ht]
    \centering
    \caption{Comparison of different models on various performance metrics. The \textit{Entangled} is an evaluation metric proposed in this paper. \textit{Entangled-D} (\textit{Entangled-Double}) is used for paired images, while \textit{Entangled-S} (\textit{Entangled-Single}) is for unpaired images, specifically evaluating the results of unlearning alone. The best values for each metric are shown in bold, and the second-best values are underlined.}
    \resizebox{1.0\linewidth}{!}{
    \begin{tabular}{ccccccccc}
    \toprule
        \textbf{Models} & \textbf{FID$\downarrow$} & \textbf{CMMD$\downarrow$} & \textbf{CLIP Dist.$\uparrow$} & \textbf{CLIP@1$\uparrow$} & \textbf{CLIP@3$\uparrow$} & \textbf{CLIP@5$\uparrow$} & \textbf{Entangled-D$\uparrow$} & \textbf{Entangled-S$\uparrow$} \\ \midrule
        \textit{Ours} & \textbf{58.8808}  & \textbf{0.8000}  & \textbf{0.9456}  & \textbf{0.8096}  & \textbf{0.7739}  & \textbf{0.6852}  & \textbf{0.8680} & \textbf{0.8119}  \\ 
        CLIPAway & \underline{68.4728}  & 0.8273  & \underline{0.8324}  & \underline{0.7493}  & 0.7027  & \underline{0.6658}  & \underline{0.7920} & \underline{0.7266}   \\ 
        LoRA+inpaint & 72.9584  & \underline{0.8231}  & 0.8211  & 0.7355  & 0.6791  & 0.5842  & 0.6997 & 0.6350 \\ 
        LoRA+I2I & 101.8833  & 0.9340  & 0.7808  & 0.7194  & \underline{0.7080}  & 0.5802  & 0.5603 & 0.5752 \\ \bottomrule
    \end{tabular}
    }
    \label{table_compare_models}
\end{table*}

\subsection{Comparison}
We validated the \textit{Entangled} metric and analyzed selective unlearning across four datasets. Our framework, incorporating both Foreground and Background components (see Figure \ref{fig_compare_models}), outperforms SOTA method CLIPAway \cite{ekin2024clipaway} in flexibility, requiring no additional training. As models improve, our framework is expected to yield even better results. Baselines LoRA+I2I and LoRA+inpaint showed reconstruction quality comparable to SOTA, confirming the dataset's relevance for selective unlearning.

Table \ref{table_compare_models} presents a quantitative evaluation of each model across various metrics, including our proposed \textit{Entangled-D} (paired images) and \textit{Entangled-S} (unpaired images) metrics, which assess image fidelity and unlearning success. Our framework achieved the best performance on nearly all metrics, with the lowest FID (58.8808) and CMMD (0.8000), and the highest \textit{Entangled-D} (0.8680) and \textit{Entangled-S} (0.8119), highlighting its superior unlearning efficacy and background consistency.
In particular, our framework achieved CLIP@1, CLIP@3, and CLIP@5 accuracies of 0.8096, 0.7739, and 0.6852, significantly outperforming CLIPAway. Its CLIP Distance reached 0.9456, about 13.6\% higher than CLIPAway, reflecting enhanced semantic consistency and object removal. For unlearning across contexts, Entangled-D and Entangled-S reliably assess paired and unpaired images, with our framework excelling in both, demonstrating its versatility and robustness.

The strong performance of LoRA+I2I and LoRA+inpaint validates the effectiveness of our framework in unlearning tasks and highlights the \textit{Entangled} metric’s adaptability and reliability for evaluating unlearning in both paired and unpaired image contexts.

\subsection{Parameters Analysis}
We investigated the impact of the LoRA scaling factor \(\alpha\) on unlearning performance, adjusting \(\alpha\) from 0.1 to 1.0 with a fixed random seed for reproducibility. Unlearning effectiveness improved as \(\alpha\) increased, with values above 0.9 showing significant unlearning effects while preserving structural consistency (see Figure \ref{fig_ablation_lora_alpha}). This trend held across datasets, with higher \(\alpha\) values enhancing separation between foreground and background.
Table \ref{table_ablation_lora_alpha} shows Entangled scores for various \(\alpha\), especially strong in the Cat dataset, where scores reached 0.8507 and 0.9080 for \(\alpha = 0.9\) and \(\alpha = 1.0\), respectively. Similar improvements were seen in the Bird and Dog datasets. These results suggest that higher \(\alpha\) values are preferable for effective selective unlearning while maintaining background integrity, with an \(\alpha\) close to 1.0 recommended for rigorous tasks.

\begin{table}[!ht]
    \centering
    \caption{Entangled scores for different LoRA scaling factors \(\alpha\).}
    \resizebox{1.0\linewidth}{!}{
    \begin{tabular}{ccccccc}
    \toprule
        $\bm{\alpha}$ & \textbf{0.1} & \textbf{0.5} & \textbf{0.7} & \textbf{0.9} & \textbf{1.0} \\ \midrule
        Bird & 0.1041  & 0.2642  & 0.5068  & 0.7727  & 0.7182  \\
        Cat & 0.1639  & 0.1469  & 0.3530  & 0.8507  & 0.9080  \\ 
        Dog & 0.1558  & 0.3011  & 0.4639  & 0.6314  & 0.6070  \\ 
        ImageNet & 0.5716  & 0.1650  & 0.1479  & 0.5849  & 0.5778  \\ \bottomrule
    \end{tabular}
    }
    \label{table_ablation_lora_alpha}
\end{table}

\section{Limitations}
\textbf{(1) Limited Handling Transparent Objects}: Compared to LayerDiffusion \cite{li2023layerdiffusion, zhang2024transparent}, our approach struggles to separate transparent objects into distinct layers, limiting its effectiveness with complex compositions.
\textbf{(2) Visual Discontinuities in Layer Merging}: When the background and foreground are not sourced from the original image, the merged result may show visual disconnects, appearing artificial due to synthesized elements that may not align with the original context and texture.
\textbf{(3) High Prompt Dependence}: The framework’s results are highly sensitive to prompt specificity. Minor prompt changes can cause the removed object to reappear, as large models can infer details even from indirect cues, reducing the robustness of the unlearning effect.

\section{Conclusion}

We propose an Automatic Dataset Creation Framework based on prompt-based layered editing and training-free local feature removal to enable selective unlearning in diffusion models, addressing the need for privacy-compliant generative models. Our contributions include a training-free removal framework for efficient dataset construction, the \textit{Entangled} metric for balanced unlearning evaluation, and the \textit{ForgetMe} dataset, a standardized benchmark with diverse real and synthetic content. Experiments validate the robustness of \textit{Entangled} and the effectiveness of LoRA-based fine-tuning. Despite challenges with transparent objects and prompt sensitivity, \textit{ForgetMe} lays a strong foundation for privacy-preserving generative AI.

{
    \small
    \bibliographystyle{ieeenat_fullname}
    \bibliography{output_v2}
}


\end{document}